# Representation in Dynamical Systems

**Matthew Hutson**
Brown University
Providence, RI
matt@silverjacket.com

### Abstract

The brain is often called a computer and likened to a Turing machine, in part because the mind can manipulate discrete symbols such as numbers. But the brain is a dynamical system, more like a Watt governor than a Turing machine. Can a dynamical system be said to operate using "representations"? This paper argues that it can, although not in the way a digital computer does. Instead, it uses phenomena best described using mathematic concepts such as chaotic attractors to stand in for aspects of the world.

## 1 Introduction

We can trace the roots of the current state of cognitive science back to Plato and his fundamental separation of mind and matter, a theory today known as substance dualism. With little science available to incorporate physical reality and experiential qualia, Plato turned away from such problems and regarded the abstract forms of the mind as the true Reality. Such a framework has directed philosophies and sciences of the mind and brain for the past 2,400 years.

Another substance dualist, Descartes, strengthened the assumptions behind the still-dominant computational paradigm. In the 17th Century he based his system of knowledge upon the self-awareness of the soul, and its own intrinsic powers of deduction. According to Descartes, "[the soul] was, after all, a *rational* soul with a scheme of innate representations, and principles of logic were the principles of its dynamics" (P.S. Churchland, 243). He understood central mental capacities merely to be processes of reasoning. According to van Gelder, "the prior grip that this Cartesian picture has on how most people think about mind and cognition makes the computational conception intuitively attractive to many people." (1995, 379).

Since the time of Aristotle, we have been able to describe logical operations explicitly and formulaically. Such operations act upon symbolic structures, which also can be described explicitly and formulaically. These abilities have led to the modern theories of computation, developed in parallel by computer scientists and cognitive scientists. Digital computers provide physical mechanisms for implementing symbolic computations, and in the computationalist paradigm the human brain is thought to be a special type of digital computer. Our knowledge and belief structures are static representations of internal or external states, and cognition takes place when abstract rules of computation are applied to such representations. Fodor (1975) calls this compositional format of cognitive computation the "Language of Thought" hypothesis. The most powerful and versatile method of representing knowledge that we know of is through symbols. Therefore, the brain must compute symbolically. A particularly well formulated version of this theory appears in Newell & Simon (1976) under the name of "the physical symbol system hypothesis." Interaction with the world takes place through peripheral input and output mechanisms that translate messy external physical states into pure symbols, and vice versa. Anything outside the range of symbol manipulation is not considered truly cognitive.

This computational framework, despite fatal flaws that we will encounter in a moment, accounts for the majority of cognitive science's success in explaining and predicting human behavior.

## 2 Problems with Computational Cognition

There are a number of theoretical and emergent difficulties with the hypothesis that human cognition is symbolic computation. The archetypal symbolic computer, the Turing machine, makes a good example for several of the theoretical issues. Firstly, such symbolic processes do not take into account time. They are described as ordered steps, but the labeled order of each step does not reveal anything about how long the step takes. In this sense, real time is arbitrary. Not only is time arbitrary, but it is also discrete. Computation does not smoothly morph one representation into another; it makes the changes instantaneously. Therefore, it does not make sense to talk about representational states in between time steps. Such temporal and physical restraints fly in the face of what we know about the real world. Therefore, strictly sequential processing is fundamentally un-brain-like in its behavior. This raises the question of how a dynamical system such as the brain would give rise to a supposedly discretely computational system such as the mind.

In addition, there exists the philosophical problem of how the physical world can effect changes in a purely mental state, and of how pure abstract information can effect changes in the surrounding physical world. This difficulty plagues any theory of mind-brain interaction (note, particularly, discussions of intentionality and mental causation) but becomes especially obtrusive when one attempts to envision the input and output mechanisms connecting a symbolic processor with the real world.

The computational hypothesis also cannot explain many emergent properties of cognition. (Due to much discussion in the literature regarding the breadth of the term "computation," I will here clarify that by computation, I mean symbolic or sentence-like computation, not connectionism, neurocomputation, etc.) The computational hypothesis holds that cognition occurs serially, one step at a time. We know cognition to be a parallel process, with many subsystems active simultaneously. We also know neural activity to be highly parallel.

Another area where the computational hypothesis has failed cognitive science is in its descriptions of human decision-making. Because of its atemporal nature, it presents no window into the process of decision-making—only the final outcome. In addition, it often fails at predicting the final outcome. Classic utility theory describes a method for calculating the best choice to make in a given situation by multiplying probabilities of consequences by their respective values. People frequently deviate from classic utility theory, hinting at nonlinear influences.

All of these difficulties set the stage for the application of dynamical systems theory.

## 3 Dynamics

What is a dynamical system? Simply, it is an element or set of elements that evolves in time. The system at time t is usually described by a state vector, a set of real, continuous numbers describing aspects of the state. Thus the state of the system, whether physical or abstract, is described geometrically as a point in space connected to an arbitrary origin by the state vector. As time increases, the state of the system changes, and the point moves through the state space. This movement is usually described using differential equations. Unlike Newtonian mechanics, where every element's interaction with every other element must be calculated continuously, the behavior of the system is usually described qualitatively as the eventual outcome of different starting points.

Some staring points will be attracted to certain areas of the state space, called attractors, or repelled from others. Sometimes they will fall into endlessly repeating cycles of movement, called limit cycles. If the cycles do not repeat exactly, but remain within certain bounds, the region is called a chaotic attractor. If a parameter is changed that alters the number or nature of these fixed points or cycles, this change is called a bifurcation. Such qualitative features of a state space can be used to envision the landscape of the system's potential behavior.

A representation is typically seen as a signal distinct from its particular physical instantiation. It's not just a force or object in and of itself, but a piece of information to be used in other contexts. To some degree, no representation instantiated in an analog (versus digital) system can be fully separated from that system (Hutson, 1998), unless by instantiating it in a physically identical substrate. But there can be approximations. A song recorded in grooves on vinyl can also be stored in electron arrangements on tape. In terms of dynamical systems, a pattern such as circular chaotic attractor can be roughly represented by a hurricane or by water circling a drain, or more cryptically by the neural correlates of an imagined tornado.

## 4 Dynamical Cognition

The math and science of dynamics has been applied fruitfully to physical and biological systems of all kinds. How does this help us explain cognition? According to van Gelder and Port, "It is the central conjecture of the Dynamical Hypothesis that [natural cognitive] systems constitute single, unified dynamical systems" (11). In this view, the behavior of a cognitive system can best be explained with differential equations describing the trajectory of the system through state space. There are too many variables involved to be counted, so the dimensionality of the system must first be reduced to simplify the description. Once a satisfactory set of variables has been established, a number of characteristically dynamical properties may appear in the cognitive system: "asymptotic approach to a fixed point, the presence or disappearance of maxima or minima, catastrophic jumps caused by small changes in control variables, oscillations, chaotic behavior, hysteresis, resistance to perturbation, and so on" (van Gelder & Port, 17). A dynamical perspective also explains several other general properties of natural cognition, such as emergent organization, change over multiple time scales, the appearance of both continuous and discrete features, and parallel processing.

Describing cognition in the same dynamical language we use to describe the brain and the environment also bridges the theoretical chasm of explaining computational processes on a dynamical machine. Now we can consider mind and world as two coupled systems naturally interacting.

## 5 Connectionism

Before we go on, I should offer special mention to the connectionist approach to cognition. Also called (artificial) neural networks, connectionist models attempt to explain adaptive behavior through the interaction of multiple computing elements, roughly modeled after neurons. These nodes are connected by "weighted" links analogous to neural synapses. The input to a node is the sum of the activity of neighboring nodes multiplied by the weights of their respective mutual links and passed through a nonlinear function. A representation is viewed as the patterns of connection weights and unit activations distributed throughout the system. This approach usually stands somewhere between computationalism and dynamicism, but some connectionists clearly stand in one camp or the other. If the input pattern is algorithmically mapped to an output pattern, then this process more closely resembles symbolic computation. There are discrete steps in the process, and discrete patterns of representation. If the process is continuous and nonlinear, it is considered dynamic, though with much higher dimensionality than traditional dynamical models. Yet the complex activity patterns can be described using simple variables and control parameters, and the overall behavior of the system can be seen to fall towards attractors or similar state space features.

## 6 The Watt Governor

We will approach the issue of representation by way of van Gelder's perennial treatment of the Watt governor, or centrifugal governor (van Gelder 1995). In the time of the industrial revolution, there existed the challenge of how to convert a steam piston's sometimes irregular oscillations into a constant angular velocity of a flywheel. The Scottish engineer James Watt designed a system by which the rotation of the flywheel drove the rotation of a vertical spindle with two metal balls attached. (See Figure 1.) As they spun faster, centrifugal force pulled them out and through an intermediary mechanism they closed the steam valve and slowed the steam-driven flywheel. Thus, under opposing forces, the wheel spun at a constant rate.

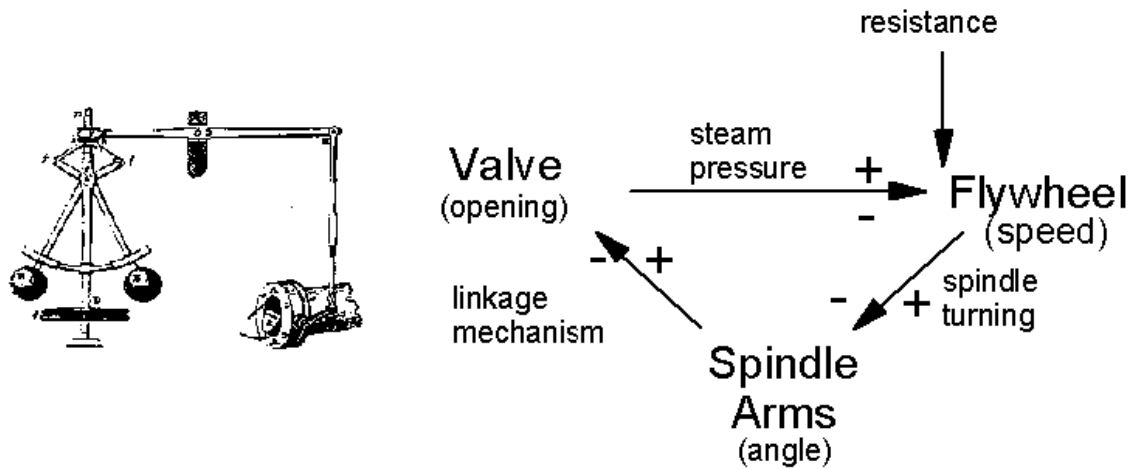

*Figure 1: The Watt governor. Bechtel, 1998.*

Van Gelder uses the Watt governor as a dynamical equivalent of the Turing machine, champion of computationalism. Specifically, he uses Watt's governor as an example of how a dynamical system can function without the use of representations; there is no system of symbols representing internal or external states. By contrast, a computational system handling the same governing problem would take a more explicit representational route:

1. Measure the speed of the flywheel.
2. Compare the actual speed against the desired speed.
3. If there is no discrepancy, return to step 1. Otherwise,
    a. measure the current steam pressure;
    b. calculate the desired alteration in steam pressure;
    c. calculate the necessary throttle valve adjustments.
4. Make the throttle valve adjustment.
Return to step 1. (van Gelder, 1995, 348.)

Computationalists propose similar homuncular, systematic algorithms as the natural way of human adaptive behavior. In fact, van Gelder states that "representation, computation, sequential and cyclic operation, and homuncularity form a mutually interdependent cluster" (351), and that each applies to the computational hypothesis. In refuting the computationalists' claims, he begins with representation.

Haugeland (1991) describes a representation as a state of a system that, according to some representational scheme used by a separate agent, stands in for some other state of the system or some state external to the system. Van Gelder acknowledged the "initially quite attractive intuition" that the angle between the vertical spindle and the arms with the swinging metallic balls represents the flywheel speed. He then offers a number of arguments against this intuition.

Firstly, he argues that if the system were representational, then it would make most sense to describe the system in representational terms. Bechtel (1998) says that it actually does make sense by pointing out that "the system is very simple, and most people see the connection [between the arm angle and the engine speed] directly" (303), thus discarding the need for an explanation such as the computational algorithm enumerated above. Despite the lack of an explicit mention of the term "representation" in one's explanation of the governor, there is an implicit acknowledgement of representational content. "For us to *understand* why this mechanism works, though, it is crucial that we understand the angle of the spindle arms as standing in for the speed of the flywheel" (Bechtel, 303).

Bechtel strikes down van Gelder's next three arguments thusly: van Gelder argues that the arm angle and engine speed are correlated but that this is not enough to designate the arm angle as a representation. Bechtel agrees, but points out that there is more than a simple correlation; a third part of the system is using the arm angle as an indicator of the engine speed. Then van Gelder says that the arm angle and engine speed are not even correlated, because there is a delay between the two. Bechtel reminds us that misrepresentation is still a form of representation. Finally, van Gelder argues that "the notion of representation is just the wrong sort of conceptual tool to apply," because the coupled relationship between arm angle and engine speed "is much more subtle and complex than the standard concept of representation can handle" (353). Bechtel's response: "the fact that the representation is in a dynamical relation with what it represents (and with the user of the representation), does not undercut its status as a representation" (304).

To the critique waged by Bechtel and others (Eliasmith, 1997; Clark & Toribo, 1994) I shall add another point. We have established the need for three elements in a representational scheme: the represented, the representer, and the user of the representation (see Figure 2). I believe the most discomfort with the representational view of the Watt governor comes from the observation that all of the interactions between elements are *forceful*. In general, the utility of a representation lies in its *informational* content. The arms do not merely carry to the steam valve information about the state of the engine—they forcefully close the valve—but this interaction belies the true nature of its function. The fact that this relationship *can* be broken down and understood as a representational one indicates the nonspecificity of its physical (forceful) nature; the abstract (representational) function of the arms could have been implemented in any of a number of imaginative ways. I propose that if the informational content of an element can be extracted from its particular physical implementation in a given system, then we may truly consider that element a representation.

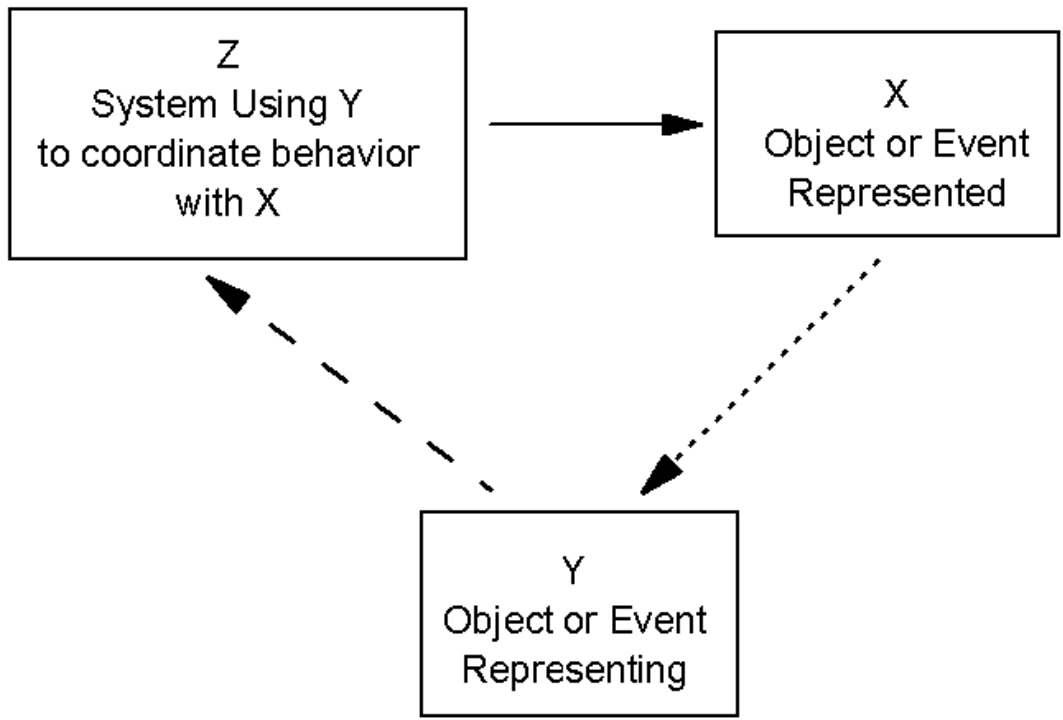

*Figure 2: The represented (X), the representer (Y), and the user of the representation (Z). Bechtel, 1998.*

## 7  The Watt Governor Extended

In the case of the Watt governor, we almost have to go out of our way to conceive of the arm angle as a representation, and one may wonder if forcefully applying the concept of representation to the realm of dynamical systems serves any purpose. Perhaps it is better just to let the systems go about their business and not read anything into them. As Brooks (1991) puts it, "when we examine very simple level intelligences we find that explicit representations and models of the world simply get in the way. It turns out to be better to use the world as its own model" (140). The field of cybernetics takes as its aim the creation of systems that simply interact directly with the world without attempting abstract descriptions of that world (Beer, 1995). In designing a cybernetic system, one traditionally focuses on perception and reaction, skipping the intermediate stage of representation. Prinz and Barsalou (2000) paraphrase Brooks's opinion (1991) on this matter: "Meeting real-world demands does not give us the luxury of representation" (70).

Does this provide an ideal model for how we should understand natural cognitive systems? Judging by the relative lengths of the stages in our evolution, most of our functionality resides in the simple ability to interact behaviorally with a changing environment. Abstract processes such as language and reasoning did not arise until much later. It turns out that for systems any more complicated than the Watt governor, representations become quite useful. Any form of robust adaptive behavior requires the ability to react to the here-and-now, as well as to predict future states of the environment and anticipate changes. This requires access to states that are not immediately available (unlike the relationship between the arm angle and engine speed in the Watt governor). The only method for accomplishing this task is the use of representation. (Beer, however, remains agnostic regarding this conclusion. He acknowledges the necessity for an internal state in any agent more than purely reactive, but he does not equate internal state with representation. At this point I feel the discussion verges toward an issue of semantics, and I thus mention Beer's hesitation only as an aside.)

Clark and Toribo (1994) point out that one of the reasons there is little utility in describing the Watt governor in representational terms is its physical nature. Human cognitive systems have the further task of "recoding gross sensory inputs so as to draw out the more abstract features to which we selectively respond" (423). Thus, state spaces do not simply describe straightforward physical attributes, but rather abstract features that are naturally couched in representational terms.

Remember our definition of representation in the previous section, however. There must be some third-party using of the representation in reference to the state represented. This implies a central executive of some kind, and the notion of a central executive doesn't sit well with the vision of a dynamical system as a free-wheeling enterprise with no one in command and everyone acting for himself. But a middle ground remains available—modular control. Dennett explained consciousness (1991) as an assembly of disparate neural/cognitive agents cooperating and competing for control over the brain/mind. Within this theory we have both bottom-up emergence and top-down executive influence acting together.

In the end, our take of the Watt governor does not displace the notion of representation from an understanding of dynamical systems, but merely encourages us to reconceive the notion of representation. A functional machine element interacting dynamically with another serves the same purpose as a set of abstract symbols arranged to describe a state of affairs. "By providing tools for analyzing how representations may change dynamically, [Dynamic Systems Theory] . . . is not challenging the use of representations but is a collaborator in understanding the format of representations" (Bechtel 1998, 305).

## 8 Dynamical Representations

"An exciting feature of the dynamical approach is that it offers opportunities for dramatically reconceiving the nature of representation in cognitive systems" (van Gelder, 1995, 376). Van Gelder hints at the generic dynamical structures that can be implemented as representations in van Gelder (1995) and van Gelder and Port (1995): state variables, parameters, states, points in state space, attractors, trajectories, bifurcations, fractal structures, "or even such exotica as transformations of attractor arrangements as a system's control parameters change." He describes a model based upon a nonlinear system perched near chaos, where regions of the state space are assigned symbolic representations, and the system's trajectory through the state space produces a sequence of symbols. In the interest of space, I will only mention a couple of examples of such dynamical representations.

Skarda and Freeman (1987) have described odor recognition as a self-organizing process that emerges out of a chaotic state. Using EEG studies of the olfactory bulbs of rabbits, they show that before an odor is introduced, the activity of the olfactory bulb is chaotic. One difference between chaos and randomness, they point out, is that chaos can be switched on an off easily through bifurcations. Each odor that an animal has learned to discriminate is represented by a limit cycle attractor. Upon sensing this odor, the olfactory bulb falls into its basin of attraction and enters the limit cycle. In addition, there is a chaotic well for dealing with novel odors. "The chaotic well provides an escape from all established attractors, so that the animal can classify an odorant as 'novel' with no greater delay than for the classification of any known sample, and it gains the freedom to maintain unstructured activity while building a new attractor" (168). Because all attractor states are easily accessible from chaos, including this chaotic well for novel stimuli, the system does not need to search through a catalog of all known odors. An additional characteristic important to note about this phenomenon is the stability of perception despite changing neural substrates. As new basins are added, the others shift slightly, yet remain intact. Essentially, "low-level context sensitivity is compatible with higher level stability" (Prinz & Barsalou, 2000, 62). This bodes well for the potential of messy dynamical systems such as the brain to implement clean computational processes.

Prinz and Barsalou (2000) lay out a number of additional claims supporting the compatibility of context sensitivity and representation. They describe models by Barsalou in which frequency, recency, and context play important roles in categorization. They explain how even prototypes and exemplars change under given demands. They also point out that forming new representations of the world relies on tracking, and tracking cannot take place without context sensitivity.

In response to the notion that cognitive representations are inherently dynamic, changing in time, some have asked how they represent parts of the world that do not change in time, such as static objects. Freyd (1987) answers this question by suggesting that we imagine stationary objects as taking part in some dynamic process. "When the perceptual system cannot directly perceive change over time it will seek out implicit evidence of change. . . future change, such as spatial transformations an object can undergo . . . [or] change that creates an object . . ." (427). She demonstrates in the paper that people apply knowledge of the muscular dynamics behind writing when perceiving handwritten letters. In this way, handwritten letters, apparently static structures, may be represented cognitively as "kinesthetic morphemes," or motor patterns.

This leads nicely into the concept of embodied cognition.

## 9 Embodied Cognition

A good introduction to the notion of embodied cognition comes by way of the symbol grounding problem. The problem, as succinctly stated by Harnad (1990): "How can the meanings of the meaningless symbol tokens, manipulated solely on the basis of their (arbitrary) shapes, be grounded in anything but other meaningless symbols?" The problem resembles one of attempting to learn Chinese using only a Chinese/Chinese dictionary. Harnad demands that the only solution can come from the bottom-up approach of grounding symbolic representations in iconic and categorical representations, both tied directly to the meaningful world of sensory perception.

Prinz and Barsalou (2000) present their theory of perceptual symbol systems as a model of cognition, based on the notion that perception and cognition share common areas of the brain and common representations. Representations are created during perception, and later they stand in for those perceptions. Many computationalists have argued that the kind of combinational, compositional semantics we see in human cognitive processes such as mental arithmetic and grammatical language perception/production requires amodal symbolic representational systems with formal syntax. Perceptual symbol systems can be both productive and propositional by combining perceptual simulations in relational ways. And, of course, the system is highly context sensitive. Context sensitivity goes a long way towards composing sophisticated cognition. "The potentially brittle and combinatorially explosive nature of general planning can be significantly alleviated by using the immediate situation to guide behavior" (Beer, 1995, 174).

Connectionists have offered many examples of parallel distributed systems learning hard-coded rules such as grammar or math. We will not discuss them here, as connectionist systems are not strictly dynamical in the sense that we have explored. However, they demonstrate that soft-coded, non-symbolic representational systems that offer the flexibility and context-sensitivity we see in human behavior can also support the symbolic processing that we see in higher-level human cognition. Thus, as researchers continue to apply dynamical concepts to human cognition, there is hope that such strategies may someday unite our disparate explanations of mental and neural functioning.

May 18, 2000